\documentclass[coverpage]{Style/inftechrep}
\usepackage{mathptmx}
\institutes{Institute for Adaptive and Neural Computation}
\reportnumber{EDI-INF-ANC-1802}
\reportyear{2018}
\reportmonth{9}

\usepackage{subcaption}
\usepackage[utf8]{inputenc} 
\usepackage[T1]{fontenc}    
\usepackage{hyperref}       
\usepackage{url}            
\usepackage{booktabs}       
\usepackage{amsfonts}       
\usepackage{nicefrac}       
\usepackage{microtype}      
\usepackage{graphicx}
\usepackage{amsmath,amssymb}
\usepackage{amsfonts}       
\usepackage{color}
\usepackage{transparent}
\usepackage{listings}
\usepackage{natbib}

\usepackage{amsmath}
\usepackage{enumerate}

\usepackage{nicefrac}       
\usepackage{footmisc}
\definecolor{dkgreen}{rgb}{0,0.6,0}
\definecolor{gray}{rgb}{0.5,0.5,0.5}
\definecolor{mauve}{rgb}{0.58,0,0.82}

\makeatletter
\newenvironment{chapquote}[2][2em]
  {\setlength{\@tempdima}{#1}%
   \def\chapquote@author{#2}%
   \parshape 1 \@tempdima \dimexpr\textwidth-2\@tempdima\relax%
   \itshape}
  {\par\normalfont\hfill--\ \chapquote@author\hspace*{\@tempdima}\par\bigskip}
\makeatother

\title{CINIC-10 Is Not ImageNet or CIFAR-10}

\author{
  Luke N. Darlow\\
  School of Informatics\\
  University of Edinburgh\\
  \texttt{l.n.darlow@sms.ed.ac.uk}
   \And
  Elliot J. Crowley\\
  School of Informatics\\
  University of Edinburgh\\
  \texttt{elliot.j.crowley@ed.ac.uk}
  \AND
     Antreas Antoniou\\
  School of Informatics\\
  University of Edinburgh\\
  \texttt{a.antoniou@sms.ac.uk} 
\And
     Amos J. Storkey\\
  School of Informatics\\
  University of Edinburgh\\
  \texttt{a.storkey@ed.ac.uk}
  
}

\begin{document}

\maketitle
 
\begin{abstract}
In this brief technical report we introduce the CINIC-10 dataset as a plug-in extended alternative for CIFAR-10. It was compiled by combining CIFAR-10 with images selected and downsampled from the ImageNet database. We present the approach to compiling the dataset, illustrate the example images for different classes, give pixel distributions for each part of the repository, and give some standard benchmarks for well known models. Details for download, usage, and compilation can be found in the associated github repository. \footnote{\label{foot:url}\url{https://github.com/BayesWatch/cinic-10}}.
\end{abstract}

\section{Motivation}

\begin{chapquote}{Anonymous Author(s)}
Recent years have seen tremendous advances in the field of deep learning~\citep{lecun2015deep}.
\end{chapquote}

Some derivation of the quote above may be familiar to many readers. Something similar appears at the beginning of numerous papers on deep learning. How might we assess statements like this? It is through benchmarking. AlexNet~\citep{alexnet} outperformed traditional computer vision methods on ImageNet~\citep{ILSVRC15}, which was in turn outperformed by VGG nets~\citep{vgg}, then ResNets~\citep{he2016deep} etc.\

ImageNet has its flaws however. It is an unwieldy dataset. The images are large, at least in neural network terms, and there are over a million of them. A single training run can take several days without abundant computational resources~\citep{imagenet1hr}. Perhaps for this reason, CIFAR-10 and CIFAR-100~\citep{cifar} have become the datasets of choice for many when initially benchmarking neural networks in the context of realistic images. Indeed, this is where several popular architectures have demonstrated their potency~\citep{dense,shakeshake}. 

In CIFAR-10, each of the 10 classes has 6,000 examples. The 100 classes of CIFAR-100 only have 600 examples each. This leads to a large gap in difficulty between these tasks; CIFAR-100 is arguably more difficult than even ImageNet. A dataset that provides another milestone with respect to task difficulty would be useful. ImageNet-32~\citep{imagenet32} already exists as a CIFAR alternative; however, this actually poses a~{\it more challenging} problem than ImageNet as the down-sampled images have substantially less capacity for information. Moreover, most benchmark datasets have uneven train/validation/test splits (validation being non-existent for CIFAR). Equally sized splits are desirable, as they give a more principled perspective of generalisation performance. 

To combat the shortcomings of existing benchmarking datasets, we present{~\bf CINIC-10: CINIC-10 Is Not ImageNet or CIFAR-10}. It is an extension of CIFAR-10 via the addition of downsampled ImageNet images. CINIC-10 has the following desirable properties:

\begin{itemize}
\item It has 270,000 images, $4.5\times$ that of CIFAR.
\item The images are the same size as in CIFAR, meaning that CINIC-10 can be used as a drop-in alternative to CIFAR-10.
\item It has equally sized train, validation, and test splits. In some experimental setups it may be that more than one training dataset is required. Nonetheless, a fair assessment of generalisation performance is enabled through equal dataset split sizes. 
\item The train and validation subsets can be combined to make a larger training set.
\item CINIC-10 consists of images from both CIFAR and ImageNet. The images from these are not necessarily identically distributed, presenting a new challenge: \emph{distribution shift}. In other words, we can find out how well models trained on CIFAR images perform on ImageNet images for the same classes.

\end{itemize}

\section{Construction}

In this section we outline our method of constructing the CINIC-10 dataset. This process is detailed with accompanying notebooks in the github repository~\citep{cinic2018github}. \footref{foot:url}

\paragraph{Stage 1: Reformatting CIFAR-10} 
The original CIFAR-10 is processed into image format (\textit{.png}) and stored as {\it set/classname/}cifar-10{\it-origin-index} where {\it set} is either train, validation or test, {\it classname} refers to the corresponding CIFAR-10 class (airplane, automobile etc.\,), {\it origin} is the set from which the image was taken (train or test), and {\it index} is the original index of the image in the set it came from. This is an equal split of the CIFAR-10 data: 20,000 images per set; 2,000 images per class within set; and an equal distribution of CIFAR-10 data among all three sets. The CINIC-10 test set contains the entirety of the original CIFAR-10 test set, as well as randomly selected examples from the CIFAR-10 train set. CINIC-10's train and validation contain a random split of the remaining CIFAR images. CIFAR-10 can be fully recovered from CINIC-10 by the filename. 

\paragraph{Stage 2: Finding relevant ImageNet images} 
The relevant synonym sets (synsets) within the Fall 2011 release of the ImageNet Database were identified and collected. These synset-groups are listed in \textit{synsets-to-cifar-10-classes.txt} in the repository. The mapping from sysnsets to CINIC-10 is listed in \textit{imagenet-contributors.csv} in the repository. These synsets were downloaded using Imagenet Utils.\footnote{\url{https://github.com/tzutalin/ImageNet_Utils}} Note that some \textit{.tar} downloads failed (with a 0 Byte download) even after repeated retries. This is not exceedingly detrimental as a subset of the downloaded images was taken.

\paragraph{Stage 3: Adding ImageNet} 
The \textit{.tar} files were extracted, the \textit{.JPEG} images were read using the Pillow Python library\footnote{\url{https://python-pillow.org/}}, and converted to $32\times 32$ colour pixel images with the Box algorithm from the Pillow library (in the same manner as ImageNet32x32~\citep{imagenet32}, for consistency). The lowest number of CIFAR10 class-relevant samples from these ImageNnet synset-groups samples was observed to be 21939 in the `truck' class. Therefore, 21000 samples were randomly selected from each synset-group to compile CINIC-10 by augmenting the CIFAR-10 data. Finally, these 21000 samples were randomly distributed (but can be recovered using the filename) within the new train, validation, and test sets, storing as follows: {\it set/classname/synsetnumber.png}, where {\it set} is either train, valid or test. {\it classname} refers to the CIFAR-10 classes (airplane, automobile, etc.). {\it synset} indicates which ImageNet synset this image came from and {\it number} is the image number directly associated with the original downloaded \textit{.jpeg} images.

This process resulted is a dataset that consists of 270000 images (60000 from the original CIFAR-10 data and the remaining from ImageNet), split into three equal-sized train, validation, and test subsets. Thus, each class within these subsets contains 9000 images.

\section{Download and Usage}
Details for download can be found at the associated repository.\footref{foot:url} The dataset is hosted on the University of Edinburgh digital repository of research data, DataShare. \footnote{\url{https://datashare.is.ed.ac.uk/}}

The simplest way to use CINIC-10 is with a PyTorch\footnote{\url{https://pytorch.org/}} data loader:

\begin{lstlisting}
import torchvision
import torchvision.transforms as transforms

cinic_directory = '/path/to/cinic/directory'
cinic_mean = [0.47889522, 0.47227842, 0.43047404]
cinic_std = [0.24205776, 0.23828046, 0.25874835]
cinic_train = torch.utils.data.DataLoader(
    torchvision.datasets.ImageFolder(cinic_directory + '/train',
    	transform=transforms.Compose([transforms.ToTensor(),
        transforms.Normalize(mean=cinic_mean,std=cinic_std)])),
    batch_size=128, shuffle=True)
\end{lstlisting}

\section{Analysis}

This section shows the difference in distribution (Section \ref{sec:distribution}) and gives examples for each class in CINIC-10 (Section \ref{sec:examples}).

\subsection{Distribution}\label{sec:distribution}

\begin{figure}[!htbp]
\centering
    \includegraphics[width=0.8\textwidth]{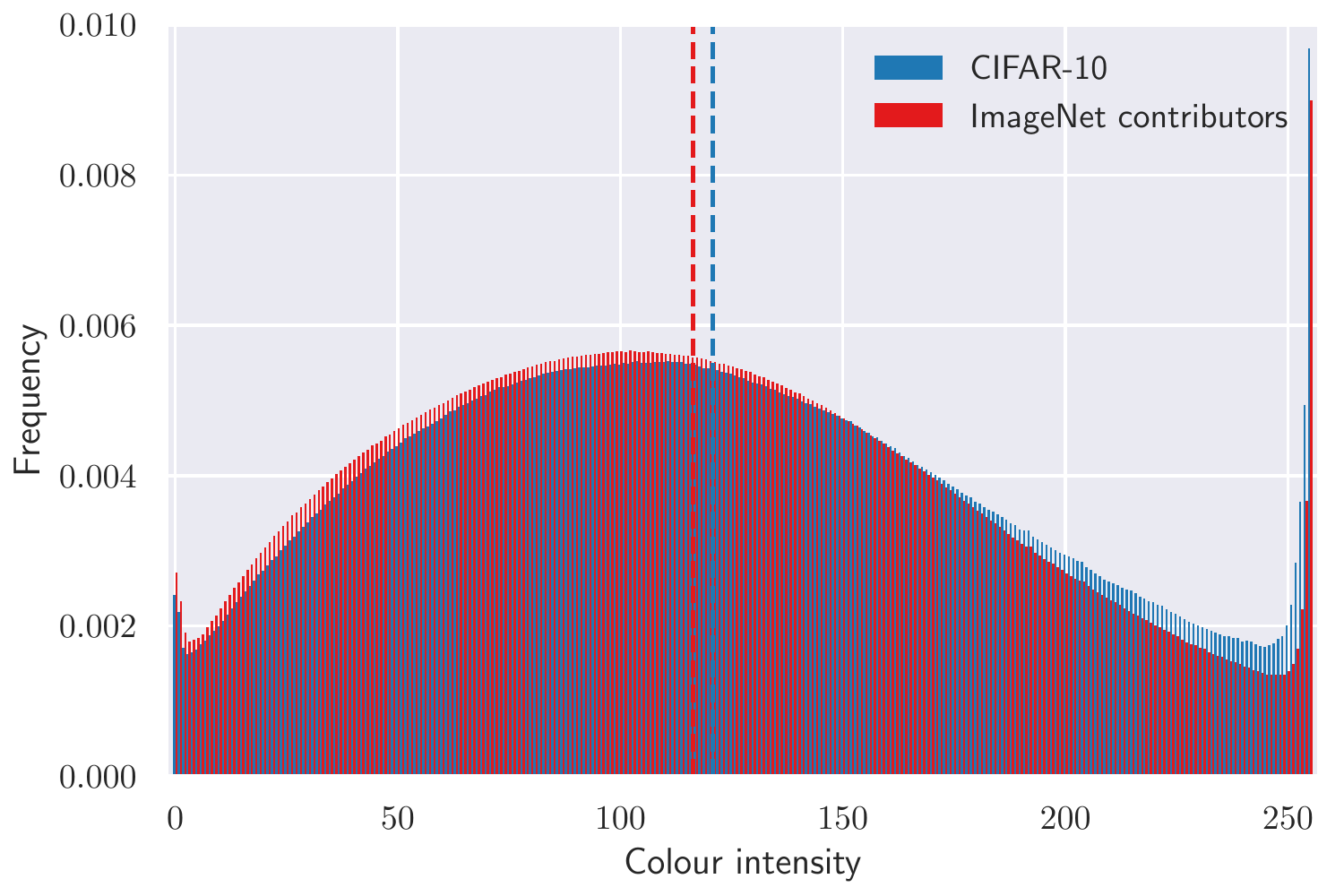}
    \caption{Histograms of the intensities (all red, green, and blue channels inclusive) of CIFAR-10 (blue) and ImageNet contributions (red). The mean is shown as the dashed lines. The colour distribution is very similar, although not identical.}
  \label{fig:distributions}
\end{figure}

The distribution of colour intensities for CIFAR-10 and ImageNet contributors is given in Figure \ref{fig:distributions}.

\subsection{Examples}\label{sec:examples}

\begin{figure}[!htbp]
\centering
    \includegraphics[width=\textwidth]{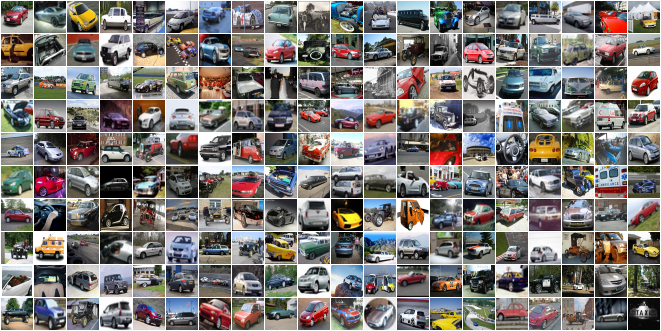}
    \caption{CINIC-10, automobile}
  \label{fig:automobile}
\end{figure}

\begin{figure}[!htbp]
\centering
    \includegraphics[width=\textwidth]{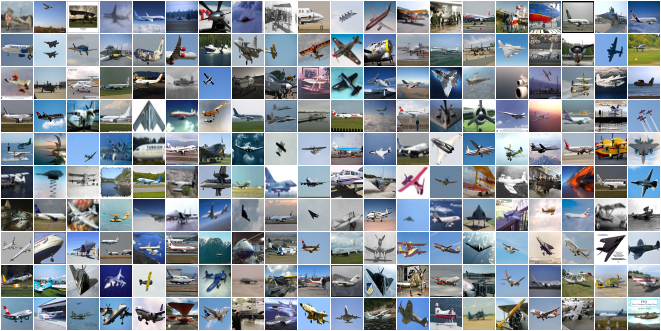}
    \caption{CINIC-10, airplane}
  \label{fig:airplane}
\end{figure}

\begin{figure}[!htbp]
\centering
    \includegraphics[width=\textwidth]{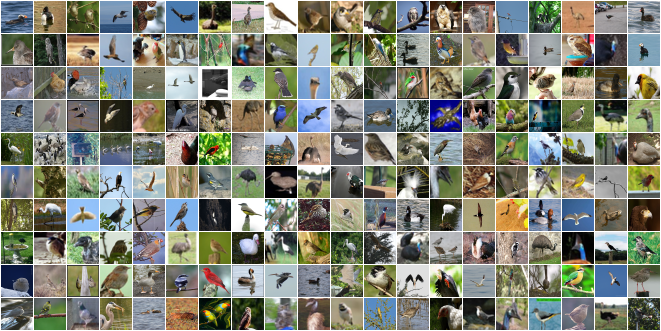}
    \caption{CINIC-10, bird}
  \label{fig:bird}
\end{figure}

\begin{figure}[!htbp]
\centering
    \includegraphics[width=\textwidth]{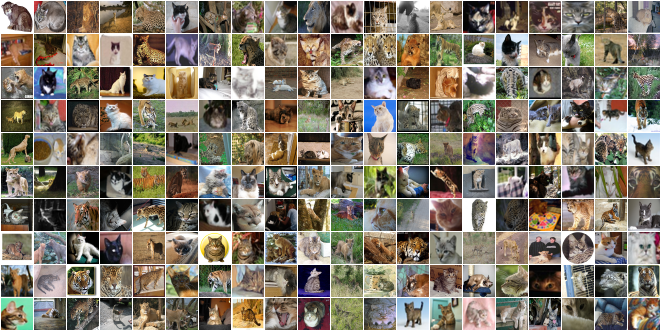}
    \caption{CINIC-10, cat}
  \label{fig:cat}
\end{figure}

\begin{figure}[!htbp]
\centering
    \includegraphics[width=\textwidth]{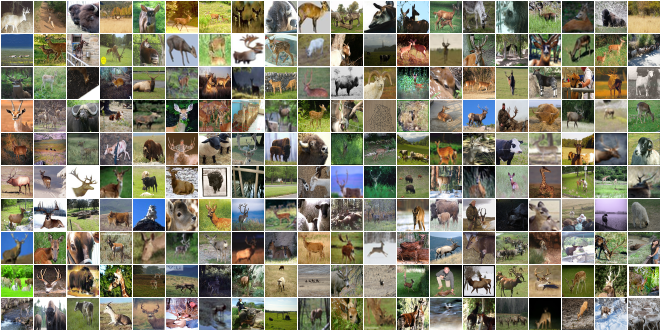}
    \caption{CINIC-10, deer}
  \label{fig:deer}
\end{figure}

\begin{figure}[!htbp]
\centering
    \includegraphics[width=\textwidth]{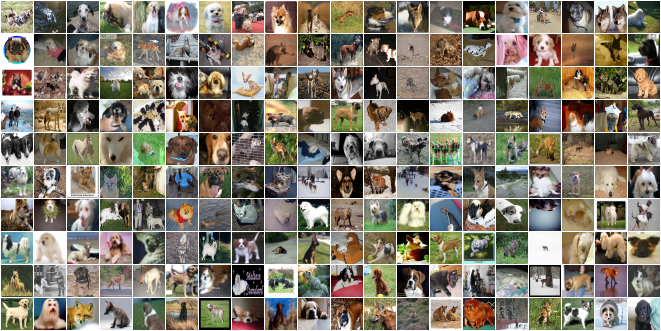}
    \caption{CINIC-10, dog}
  \label{fig:dog}
\end{figure}

\begin{figure}[!htbp]
\centering
    \includegraphics[width=\textwidth]{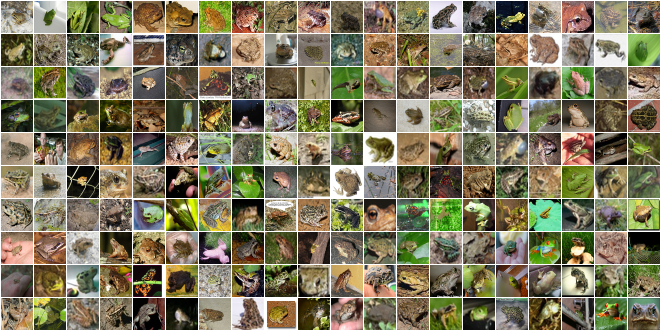}
    \caption{CINIC-10, frog}
  \label{fig:frog}
\end{figure}

\begin{figure}[!htbp]
\centering
    \includegraphics[width=\textwidth]{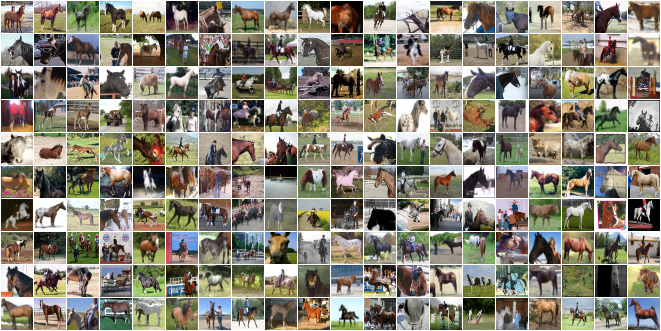}
    \caption{CINIC-10, horse}
  \label{fig:horse}
\end{figure}

\begin{figure}[!htbp]
\centering
    \includegraphics[width=\textwidth]{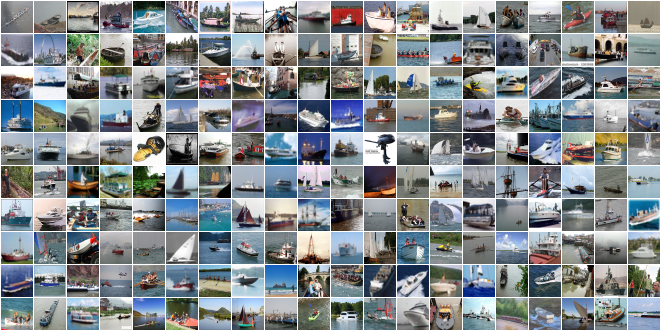}
    \caption{CINIC-10, ship}
  \label{fig:ship}
\end{figure}

\begin{figure}[!htbp]
\centering
    \includegraphics[width=\textwidth]{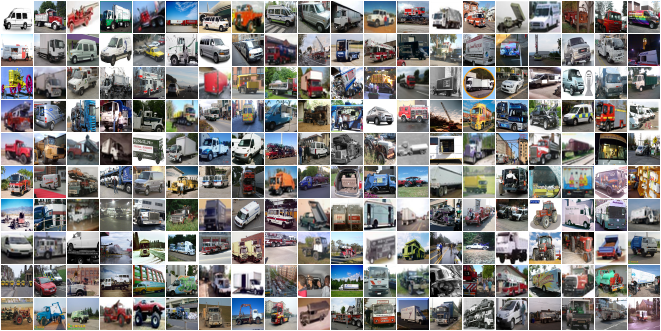}
    \caption{CINIC-10, truck}
  \label{fig:truck}
\end{figure}

Figures \ref{fig:automobile} through \ref{fig:truck} show randomly selected samples from CINIC-10. Readers familiar with CIFAR-10 images will note that these are more noisy, not necessarily explicitly cropped and centred, and may contain elements that are less class-distinct as CIFAR-10 (cows and goats in the deer class, for instance).

\section{Benchmarks}

Table \ref{tab:bench} gives benchmark results on CINIC-10. Model definitions that were applied to CIFAR-10\footnote{\url{https://github.com/kuangliu/pytorch-cifar/}} were copied and tested. 

\begin{table}[!htbp]
\def\arraystretch{1.5}
\caption{CINIC-10 benchmarks.}\label{tab:bench}
\centering
\begin{tabular}{lrr}
\hline
\textbf{Model}              & \textbf{No. Parameters} & \textbf{Test Error} \\
\hline
VGG-16             & 14.7M          & 12.23 $\pm$ 0.16   \\
ResNet-18          & 11.2M          & 9.73 $\pm$ 0.05    \\
GoogLeNet          & 6.2M           & 8.83 $\pm$ 0.12    \\
ResNeXt29\_2x64d   & 9.2M           & 8.55 $\pm$ 0.15    \\
DenseNet-121       & 7.0M           & 8.74 $\pm$ 0.16    \\
MobileNet          & 3.2M           & 18.00 $\pm$ 0.16   \\
\hline
\end{tabular}

\end{table}

\section{Conclusion}
We presented CINIC-10 in this technical report. It was compiled by augmenting CIFAR-10 with downsampled images sourced from ImageNet. We proposed that benchmarking is paramount in machine learning and that there is a blind-spot regarding dataset size and task difficulty. To this end we offer a new benchmarking dataset that is larger than CIFAR-10 (and more challenging) but not as difficult as ImageNet.

\bibliography{iclr2019_conference}
\bibliographystyle{iclr2019_conference}

\end{document}